\documentclass{article}

\usepackage{arxiv}

\usepackage[utf8]{inputenc} % allow utf-8 input
\usepackage[T1]{fontenc}    % use 8-bit T1 fonts
\usepackage{hyperref}       % hyperlinks
\usepackage{url}            % simple URL typesetting
\usepackage{booktabs}       % professional-quality tables
\usepackage{amsfonts}       % blackboard math symbols
\usepackage{nicefrac}       % compact symbols for 1/2, etc.
\usepackage{microtype}      % microtypography
\usepackage{lipsum}		% Can be removed after putting your text content
\usepackage{graphicx}
\usepackage{natbib}
\usepackage{doi}
\usepackage{makecell}
\usepackage{amsmath}
\usepackage{caption}

\renewcommand{\undertitle}{Submitted to The 36th International Conference on Principles of Diagnosis and Resilient Systems (DX'25)}

\title{
Optimized Spectral Fault Receptive Fields for Diagnosis-Informed Prognosis}

\date{}

\author{ \href{https://orcid.org/0000-0001-6259-1609}{\includegraphics[scale=0.06]{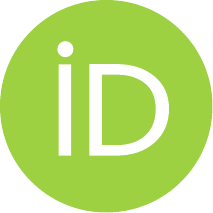}\hspace{1mm}Stan Muñoz Guitiérrez}\thanks{Authors are listed in alphabetical order.} \\
	Institute of Software Engineering and\\ Artificial Intelligence\\
    Graz University of Technology \\
    Münzgrabenstraße 137, A-8010 Graz, Austria \\
	\texttt{munozgutierrez@tugraz.at} \\
	\And
	\href{https://orcid.org/0000-0002-0462-2283}{\includegraphics[scale=0.06]{orcid.pdf}\hspace{1mm}Franz Wotawa}\thanks{Corresponding author} \\
	Institute of Software Engineering and\\ Artificial Intelligence \\
    Graz University of Technology \\
    Inffeldgasse 16b/2, A-8010 Graz, Austria \\
	\texttt{wotawa@tugraz.at} \\
}

\hypersetup{
pdftitle={Optimized Spectral Fault Receptive Fields for Diagnosis-Informed Prognosis},
pdfsubject={cs.AI},
pdfauthor={Stan Muñoz Guitiérrez, Franz Wotawa},
pdfkeywords={Health Perception,
Spectral Fault Receptive Fields,
Remaining Useful Life,
Incipient Fault Diagnosis,
Prognostics and Health Management,
Condition Monitoring,
Evolutionary Multi-Objective Optimization,
Bagged Regression Tree Ensemble,
Bearing Fault Diagnosis},
}

\begin{document}

\maketitle

\begin{abstract}
This paper introduces Spectral Fault Receptive Fields (SFRFs), a biologically inspired technique for degradation state assessment in bearing fault diagnosis and remaining useful life (RUL) estimation.
Drawing on the center-surround organization of retinal ganglion cell receptive fields, we propose a frequency-domain feature extraction algorithm that enhances the detection of fault signatures in vibration signals.
SFRFs are designed as antagonistic spectral filters centered on characteristic fault frequencies, with inhibitory surrounds that enable robust characterization of incipient faults under variable operating conditions.
A multi-objective evolutionary optimization strategy based on NSGA-II algorithm is employed to tune the receptive field parameters by simultaneously minimizing RUL prediction error, maximizing feature monotonicity, and promoting smooth degradation trajectories.
The method is demonstrated on the XJTU-SY bearing run-to-failure dataset, confirming its suitability for constructing condition indicators in health monitoring applications.
Key contributions include: (i) the introduction of SFRFs, inspired by the biology of vision in the primate retina; 
(ii) an evolutionary optimization framework guided by condition monitoring and prognosis criteria; and 
(iii) experimental evidence supporting the detection of early-stage faults and their precursors. 
Furthermore, we confirm that our diagnosis-informed spectral representation achieves accurate RUL prediction using a bagging regressor.
The results highlight the interpretability and principled design of SFRFs, bridging signal processing, biological sensing principles, and data-driven prognostics in rotating machinery.

\end{abstract}

\section{Introduction}

% \begin{figure}[!htbp]
%     \centering
%     \includegraphics[width=0.8\linewidth]{figures/reliability.pdf}
%     \caption{Visual depiction of reliability definitions from several engineering domains and international standards.}
%     \label{fig:reliability-standards}
% \end{figure}

In modern engineering, reliability is a central concern, distinguished from quality by its emphasis not only on compliance with specifications at ``time zero,'' but also on sustained performance throughout an artifact’s operational life. 
Central to reliability is the assessment and modeling of degradation rates and time to failure~\cite{mcpherson:2019}.
% Figure~\ref{fig:reliability-standards} visually compares reliability definitions across various engineering domains and international standards. 
% According to these standards, 
Reliability pertains to the performance and operation of systems and their components, aiming to deliver solutions that can operate without failure, nor be the cause of failure, over a specified time horizon and in accordance with specifications that define both constraints and operational conditions.

Rotary machines are ubiquitous in industrial and transportation contexts. Bearings, as key components of these machines, play a crucial role in ensuring reliable operation. Accurately estimating the degradation state of bearings throughout their operational life is essential for rational decision-making by both humans and automated systems. Such decisions include scheduling maintenance actions, investigating accelerated degradation trends, predicting component failure times, implementing closed-loop control for safety and energy efficiency, and, when done effectively, extending component lifetimes through feedback-driven control based on degradation state estimation.

As part of Project Archimedes, we are investigating intelligent, data-driven approaches for accurate degradation state estimation and remaining useful life prediction, with the goal of enabling decisions that extend the operational life of electric vehicle powertrains (EVPs). An important challenge in this line of research is that knowledge on bearing diagnostics is mostly available for bearings operating under constant operational conditions, which are unrealistic for electric drives in vehicles, where speed and load change dynamically in the presence of disturbances and aleatoric sources of uncertainty. 
Although we do not address this aspect directly in our research, our solution is designed to incorporate such parameters, and we briefly discuss their implications. 
Another important challenge is the scarcity of open data; currently, there are only a few open datasets for electric drive run-to-failure components, which severely hinders research in this field. For this study, we rely on the XJTU-SY dataset~\cite{wang:lei:li:li:2020}, one of the few exceptions to this rule.

The work presented in this paper focuses on bearings, which are integral to the mechanisms that connect the electric drive to the transmission and enable vehicle propulsion. 
However, we have developed our model to be generalizable to some extent, making it potentially applicable to other aspects of the electric drive, such as electric winding faults, irreversible demagnetization of permanent magnets, and inverter degradation dynamics, assuming suitable adaptations are implemented.

Our work introduces a novel technique based on consolidated knowledge within the field of vibration analysis. 
Although many recent research efforts adopt tabula rasa methodologies, bypassing established domain knowledge in favor of black-box systems that often achieve high performance, these solutions frequently lack transparency and interpretability. In safety-critical domains such as electric drives, transparency is essential; certification requires that system behavior be understandable and trustworthy.

Our method, named Spectral Fault Receptive Fields, offers an interpretable technique to degradation state estimation, with condition indicators that correspond directly to specific failure modes in bearings. 
We evaluated the system primarily using the monotonicity criterion, and further incorporated smoothness and remaining useful life (RUL)-based metrics for parameter selection. 
Through qualitative comparison, we demonstrated clear improvements resulting from explicit multi-objective optimization of several system parameters, thereby validating the effectiveness of the approach.

\section{Related Work}

Traditionally, reliability assessments were based primarily on empirical field data derived statistically~\cite{mil-hdbk-217:1991,mcleish:2010}. These approaches were typically static, focused solely on random failures, often neglected underlying failure mechanisms, did not account for differences among vendors or specific devices in lifetime predictions, and excluded real-time condition monitoring~\cite{mcleish:2010}.
In contrast, modern diagnostics and prognostics frameworks emphasize continuous degradation monitoring of components and systems. They employ a range of modeling strategies, including failure progression rates, physics-of-failure, statistical and probabilistic methods, and modeling of failure propagation between interconnected subsystems. This entails a more comprehensive and dynamic assessment of system health~\cite{vachtsevanos:lewis:roemer:hess:wu:2006}.
When the primary focus is on selecting optimal maintenance actions, predictive maintenance (PdM) serves as an appropriate conceptual framework. However, research in this area typically concentrates on two aspects, which are seldom addressed simultaneously: (1) predicting the time to failure, referred to as remaining useful life (RUL) prediction, and (2) optimizing maintenance strategies.
Prognostics and health management (PHM) is conducive to informed decision-making and actions to keep systems in optimal operating condition. PHM is an integrated, modular process that includes system analysis, data acquisition, data processing, fault detection, fault diagnostics, failure prognostics, decision making, and maintenance scheduling~\cite{soualhi:nguyen:medjaher:nejjari:puig:blesa:quevedo:marlasca:2023}.
A typical predictive maintenance workflow consists of the following steps: (1) data acquisition and organization, (2) data preprocessing, (3) development of a fault detection or prediction model, and (4) deployment and integration~\cite{mathworks:gettingstarted:2024}. In this study, we focus primarily on step (3), which practitioners often divide into two sub-tasks: (i) the design of condition indicators and (ii) model training for fault detection or prediction tasks.
The design of \emph{condition indicators} (CIs) encompasses the computation and selection of features that correlate with the state of health of the system. A \emph{health indicator} (HI) combines multiple condition indicators into a single efficient indicator that is highly informative of degradation~\cite{mathworks:gettingstarted:2024,nguyen:dieulle:grall:2015}.
The separation of sub-tasks (i) and (ii) is instrumental in tackling the complexity of the problem, but often leads to suboptimality or extensive iterative improvements. 
We will address this concern in our contribution by means of multi-objective optimization methods that can inform the HI design, factoring in its prognostic efficacy.

For bearings, degradation is irreversible. 
Tracking the degradation state throughout the component’s operational life can be effectively achieved with suitable sensors and signal processing techniques. 
The most prevalent failure mechanism under nominal conditions (where bearings are correctly installed and lubricated) is subsurface-originated spalling, which can be detected at an early stage using acoustic emission sensors\cite{hidle:2021}.
Oil analysis sensors are highly effective for early detection of degradation in bearings and gearboxes, as they quantify the accumulation of debris from the onset of wear processes\cite{gill:weardetect:2024}. 
While primarily limited to surface-related defects, the use of MEMS-based accelerometer sensors in our application domain enables cost-effective health monitoring solutions.
Consequently, there is strong research interest in developing representations and algorithms capable of capturing early degradation and detecting incipient faults.
Accelerometers are the standard transducers for helicopter gearbox condition monitoring, providing essential input to health and usage monitoring (HUM) systems. 
A healthy transmission exhibits a characteristic \emph{fingerprint}, referred to as the \emph{regular meshing components of the signal}~\cite{samuel:pines:2005}. 
For rolling bearings, the characteristic frequencies of their components are well established~\cite{randall:2004,shi:wang:qu:2004}, and their computation is readily available in standard predictive maintenance solutions~\cite{mathworks:usersguide:2024}. Our work leverages this domain knowledge by explicitly computing representations that focus on the characteristic frequencies of bearing elements, aligning with established practices in vibration analysis and intelligent fault diagnosis for rotating machinery.

Faults can be classified based on their severity into three main categories: (1) abrupt, also known as stepwise fault; (2) incipient, also known as drifting; and (3) intermittent~\cite{park:fan:hsu:2020}. \emph{Incipient faults} in bearings have weak signal signatures that are difficult to detect due to their stochastic nature, multiple transmission paths, and the aleatoric uncertainty present in mechanical systems~\cite{hu:zhang:liang:wang:2009}.
Because of this, despite their well-understood spectral signatures, incipient fault detection remains an active area of research. 
Vibration signal analysis and modeling typically utilize degradation models that define at least two primary stages: (1) a flat, horizontal region corresponding to the healthy state, where remaining useful life (RUL) prediction is generally unreliable and arguably unnecessary, and (2) a degraded unhealthy state, which is the main focus of most analytical techniques for delivering accurate RUL estimates~\cite{lei:li:guo:li:yan:lin:2018}. 
Piecewise linear models are often used to model this change in degradation dynamics~\cite{qin:yang:zhou:pu:mao:2023}.
Current research in bearing condition monitoring and prognosis is increasingly focused on extending the prediction horizon to encompass as much of the component's operational life as possible.
In our present paper, we devise biologically inspired condition indicators that address the characterization of early degradation stages and not only correlate with manifested abrupt abnormalities.

There is a wide availability of vibration-based condition indicators in the literature. 
A taxonomy by Yan et al.~\cite{yan:qiu:iyer:2008} classifies them according to the representation domain: (1) time domain, (2) frequency domain, (3) time-frequency, and (4) wavelet. 
While this classification is not exhaustive, excluding some nonlinear feature extraction methods such as chaos-theoretic-based~\cite{hu:zhang:liang:wang:2009} and information-theoretic-based~\cite{shi:wang:qu:2004,caesarendra:kosasih:tieu:moodie:2013}, it nevertheless effectively represents the dominant approaches in the field.
Among the most widely adopted characterizations are two statistical properties that can be computed regardless of the representational domain: (1) kurtosis-based, often spectral, and (2) RMS-based (with safety and vibration severity assessed by this metric, as in ISO 10816~\cite{iso20816-3:2022}). 
Both are effective and can be used complementarily for different stages of the degradation process~\cite{engel:gilmartin:bongort:hess:2000}, while proven effective across diagnostic~\cite{moshrefzadeh:fasana:2018}, condition monitoring~\cite{meng:yan:chen:liu:wu:2021}, or prognostic~\cite{medjaher:tobon-mejia:zerhouni:2012} tasks. 
Our contribution builds on the frequency spectrum of vibration signals and is specifically designed for a low computational footprint, ensuring that it does not add significantly to the computational cost of the fast Fourier transform (FFT).

Our primary objective in this work is to engineer \emph{health perception} systems capable of actively tracking the degradation state of bearings in alignment with defined cost and reliability constraints, thereby enabling accurate estimation of RUL.
We adopt the term perception to underscore the biological inspiration behind our approach to CI construction.
In biological systems, effective perception, of both the self and the environment, is essential to survival: proprioception, homeostatic regulation~\cite{asaro:2008}, and autonomic functions support internal integrity and health, while environmental perception enables adaptive responses.
Our approach subscribes to the principles of autopoiesis~\cite{maturana:varela:1972} and biological autonomy~\cite{moreno:mossio:2015}, where integrity is understood as an emergent property of systemic organization and constraint equilibria.
This aligns well with closed-loop lifetime and degradation control algorithms~\cite{felix:Martinez:berenguer:2024}.

To guide the construction of CIs, we draw inspiration from the biology of vision, specifically, the theory of center-surround opponency in the trichromatic visual system of primates, and adopt an adapted version of the difference-of-Gaussians (DoG) model, widely applied in both biological and artificial vision domains~\cite{somaratna:freeman:2025}.
Derived from a novel transfer of computational models, we faced in our work the problem of appropriate parameterization of our Difference of Gaussians (DoG) method. Selecting the spectro-spatial scales relevant to bearing faults was achieved by relying on engineering judgement informed by field experience, a process we refer to as empirical parameter selection. 
We provide evidence that a DoG configured with these empirically chosen parameters encodes CIs effectively.
To refine the model further for predictive-maintenance and prognostics applications, we optimised its parameters with a multiobjective genetic algorithm~\cite{deb:pratap:agarwal:meyarivan:2002} guided by established condition-monitoring and prognosis criteria.
Criteria must be quantifiable and provide foundations for certification~\cite{saxena:celaya:balaban:goebel:saha:saha:schwabacher:2008}.
Our results reveal a tradeoff, present among local Pareto-optimal front members, between the monotonicity criterion, widely advocated for health indicators \cite{engel:gilmartin:bongort:hess:2000,qin:zhang:hu:sun:he:lin:2017,chen:li:wang:liu:2024,fu:kwon:huh:liu:2025}, and the accuracy of remaining-useful-life (RUL) predictions measured via normalised mean-squared error (MSE).
Related work by Qin et al.~\cite{qin:zhang:hu:sun:he:lin:2017} employs genetic programming to evolve an arithmetic condition indicator optimised for monotonicity within a Wiener stochastic framework that is subsequently refined through expectation-maximisation.

\section{Problem Domain and Dataset}
\subsection{XJTU-SY Bearing dataset}

The experimental evaluation in this work is based on the XJTU-SY\cite{wang:lei:li:li:2020} run-to-failure bearing dataset, which provides high-resolution vibration data for multiple bearings subjected to different operating conditions until failure. The testbed comprises several groups of bearings, each operated at a fixed rotational speed and load, with vibration signals acquired using both \emph{horizontal} and \emph{vertical} acceleration sensors.
Knowledge of the bearing's physical parameters is essential for accurate health monitoring and fault diagnosis. All bearings used in the dataset are heavy-duty LDK UER204 model, relevant parameters are summarized in Table~\ref{tab:bearing-parameters}.

\begin{table}[ht]
\centering
\begin{tabular}{lclc}
\toprule
\textbf{Parameter}      & \textbf{Symbol} & \textbf{Value} & \textbf{Units} \\
\midrule
Inner raceway diameter  &     $D_I$                      & 29.30  & mm  \\
Outer raceway diameter  &     $D_O$                      & 39.80  & mm  \\
Pitch diameter          &     $D_P$                      & 34.55  & mm  \\
Ball diameter           &     $D_B$                      & 7.92   & mm  \\
Contact angle           &     $\phi$                     & 0      & deg \\
\bottomrule
\end{tabular}
\caption{Relevant design parameters of the bearing LDK UER204.}
\label{tab:bearing-parameters}
\end{table}

Each bearing is uniquely identified by a label (e.g., ``Bearing1\_1''), and the dataset is organized into snapshots, each representing a short time window of recorded vibration data. 

\subsection{Vibration Signatures of Bearings}
\label{sec:vibration-signatures-bearings}
We are interested in monitoring the degradation of the different elements of a bearing to detect incipient faults. Surface defects in these elements produce well-understood vibration signatures at characteristic frequencies, determined by the bearing’s geometry and operational speed. 
%When rolling elements interact with these defects, they generate impulsive signals primarily in frequency bands that depend on the  defect location, bearing design parameters, speed, and load.

\begin{figure}[htb]
\centering
\begin{minipage}[t]{0.52\textwidth}
    \vspace{0pt}
    \begin{tabular}{ll}
        \toprule
        \textbf{Acronym} & \textbf{Equation} \\
        \midrule
        BPFO & $f_{\mathrm{BPFO}} =  f_r \frac{N_B}{2} \left[1-\frac{D_B}{D_P}\cos{\phi}\right]]$ \\
        BPFI & $f_{\mathrm{BPFI}} =  f_r \frac{N_B}{2} \left[1+\frac{D_B}{D_P}\cos{\phi}\right]$ \\
        BSF  & $f_{\mathrm{BSF}} = f_r  \frac{D_P}{2 D_B} \left[1-\left(\frac{D_B}{D_P}\cos{\phi}\right)^2\right]$ \\
        FTF  & $f_{\mathrm{FTF}} = \frac{f_r }{2} \left[1-\frac{D_B}{D_P}\cos{\phi}\right]$ \\
        \bottomrule
    \end{tabular}
    \captionof{table}{Characteristic frequencies related to bearing faults. BPFO = Ball Pass Frequency Outer Race; BPFI = Ball Pass Frequency Inner Race; BSF = Ball Spin Frequency; FTF = Fundamental Train Frequency (Cage).
    $N_B$: number of rolling elements; $D_B$: ball diameter; $D_p= \frac{D_I+D_O}{2}$: pitch diameter ; $\phi$: contact angle; $f_r$: shaft rotational frequency.}
\end{minipage}
\hfill
\begin{minipage}[t]{0.42\textwidth}
    \vspace{0pt}
    \centering
    \includegraphics[width=0.85\textwidth]{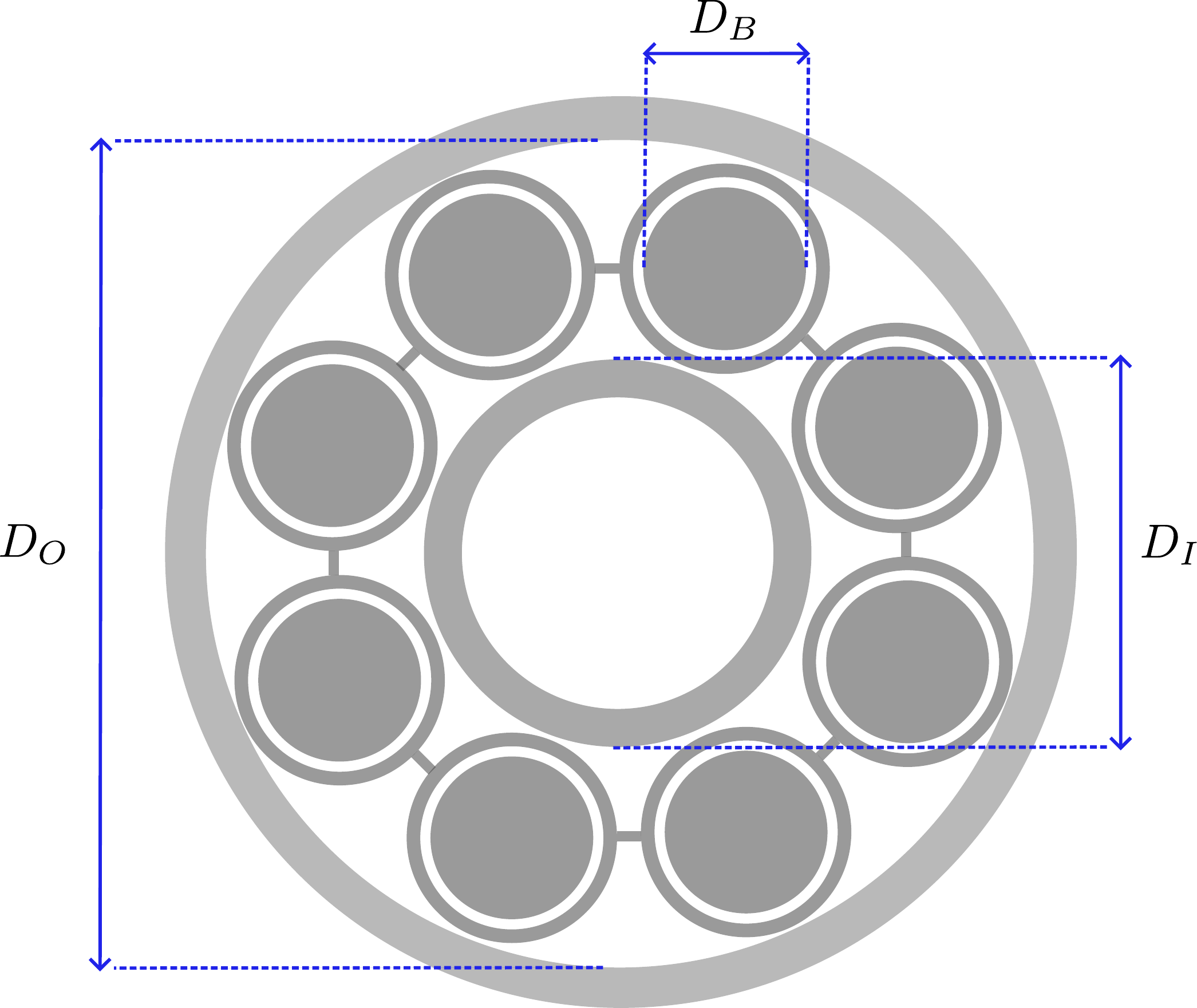}
    \captionof{figure}{Schematic diagram of bearing geometry and parameters.}
\end{minipage}
\end{figure}

\begin{figure}[htb]
\centering
\begin{minipage}[t]{0.48\textwidth}
    \vspace{0pt}
    \centering
    \includegraphics[width=0.85\textwidth]{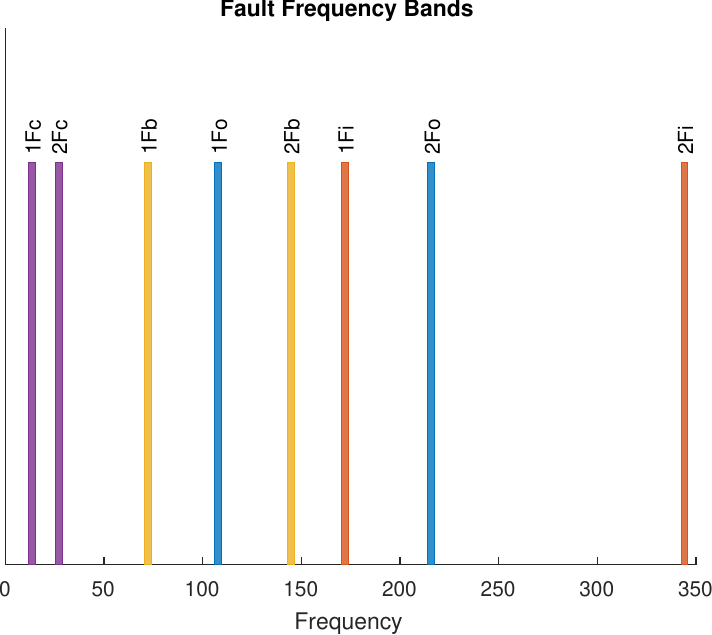}
    \captionof{figure}{Fault Frequency Bands for the first and second harmonics. Notation: nF: the n-th harmonic for frequency F, $F\in \{\text{Fo},\text{Fi},\text{Fc},\text{Fb}\}$. Frequencies are Fo: BPFO ( $f_{\mathrm{BPFO}} = 107.9074$), Fi: BPFI ($f_{\mathrm{BPFI}} = 172.0926$), Fc : FTF ($f_{\mathrm{FTF}} = 13.4884 Hz$), and Fb: BSF ($f_{\mathrm{BSF}} =72.3300$)}
    \label{fig:harmonics}
\end{minipage}
\hfill
\begin{minipage}[t]{0.48\textwidth}
    \vspace{0pt}
    \centering
    \includegraphics[width=0.85\textwidth]{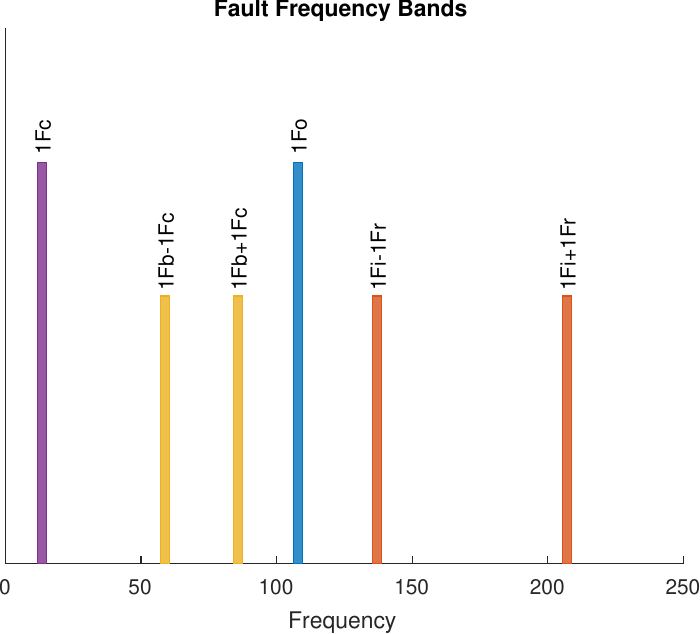}
    \captionof{figure}{Fault Frequency Sidebands. Notation: nFb-mFc: the m-th negative sideband of n-th (central) harmonic frequency of Fb (nFb-mFc is obtained as $n \times F_b - m \times F_c$), nFb+mFc: the m-th positive sideband of n-th (central) harmonic frequency of Fb (obtained with the sum).}
    \label{fig:sidebands}
\end{minipage}
\end{figure}

%\subsubsection{Frequency Bands and Sidebands}
%\label{sec:frequency-bands}
Vibration signature analysis fundamentally depends on monitoring changes in vibration near the characteristic frequencies of bearings. 
As degradation progresses, the activity within these frequency bands evolves, reflecting the bearing’s health state. 
Building on this established principle, our processing pipeline begins by computing these characteristic bands. 
While increased excitation in these bands is a hallmark of bearing defects, such activity can also be present throughout the bearing’s operational life. 
For clarity and brevity, we refer to these as \emph{fault frequency bands}, though their activity is not exclusively associated with faulty conditions, as some excitation is present throughout the bearing’s operational life. Figure~\ref{fig:harmonics} shows the default sidebands obtained by MATLAB.

%\subsubsection{Frequency Sidebands}
%\label{sec:frequency-sidebands}
Amplitude modulation is a common phenomenon in bearing vibration signals, particularly in the presence of certain faults. For inner race defects, the fault interacts with the shaft’s rotational speed; because the load distribution varies during each rotation, this results in modulation of amplitude where the characteristic frequency of the inner race fault acts as a carrier and the shaft rotational frequency serves as a modulating frequency. Similarly, amplitude modulation can occur between the ball spin frequency (BSF) and the fundamental train frequency (FTF), with the BSF as the carrier. This modulation occurs as the ball moves in and out of the load zone during cage rotation. Figure~\ref{fig:sidebands}, shows the first-order sidebands for these phenomena.

\section{Spectral Fault Receptive Fields}
\begin{figure}[t]
    \centering
    \includegraphics[width=0.6\linewidth]{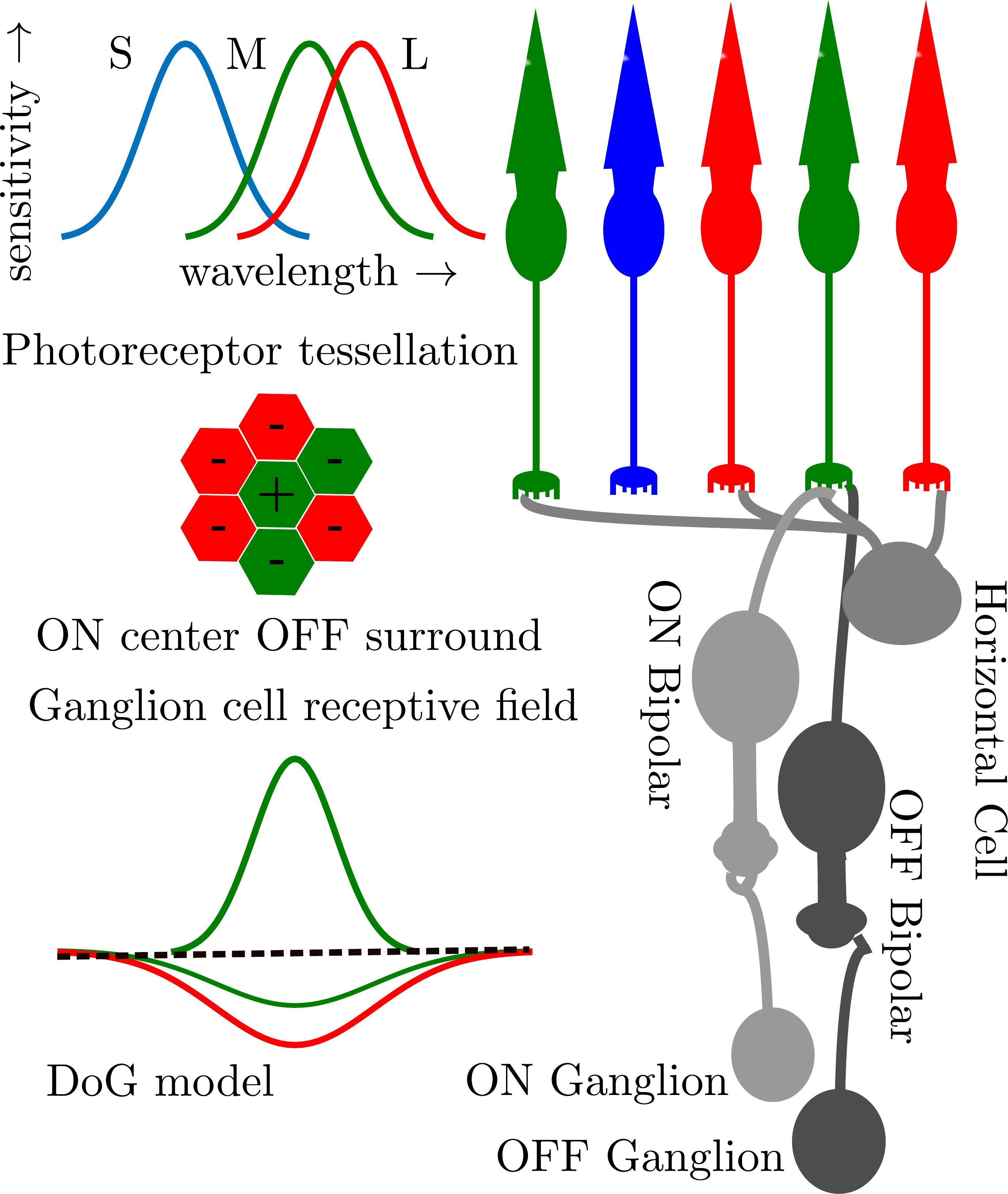}
    \caption{Encoding pathway underlying receptive field formation in primate retinal midget (P) ganglion cells (with apologies to Ramon y Cajal). Many intermediate cell types have been abstracted away from the diagram for clarity.
    Spectral sensitivities of photopigments in cone photoreceptors are shown at the top left: S-cones respond to short wavelengths, M-cones to middle wavelengths, and L-cones to long wavelengths. 
    Acute vision arises from the computation of spatial and chromatic contrast through interactions among photoreceptors, bipolar cells, and horizontal cells. 
    Bipolar cells function as ON or OFF channels depending on glutamate regulation and receptor types at cone-bipolar synapses, responding to light increments or decrements, respectively. 
    Surround inhibition mediated by horizontal cells sharpens contrast via center-surround antagonism, a process commonly modeled by the difference of Gaussians (DoG) model in biological and artificial vision systems.}
    \label{fig:ganglion-receptive-fields}
\end{figure}

For each one of the faults, we will construct fault detectors inspired by the primate retinal ganglion cell receptive fields. 
Receptive fields in the primate retina are specific regions of the visual field where the presence of a stimulus (such as light or its absence) can excite (or inhibit) the activity of a ganglion cell. 
Although the retina encodes visual information through many parallel channels (processing chromatic, spatial, and temporal information), many image-forming retinal ganglion cells share a fundamental property: a center-surround spatial and chromatic organization. 
This type of processing encodes information about spatial and for some cells also chromatic contrast or differential excitation within the receptive field's spatial extent.
Figure~\ref{fig:ganglion-receptive-fields} illustrates the classical view of the processing pathway leading to the receptive field formation of a midget ganglion cell, which computes chromatic and contrast in the red-green channel.

\subsection{Frequency Filters}
To define the fault detectors, we utilize the frequency bands described in Section~\ref{sec:vibration-signatures-bearings} %and~\ref{sec:frequency-sidebands}. 
Drawing inspiration from the center-surround organization of ganglion cell receptive fields, we introduce two distinct \emph{spectral extents}, that is, frequency bands, with a narrower band representing the center and a broader band representing the surround. 
We adopt a center-to-surround bandwidth ratio of 1:3, with the central bandwidth set to 1/10 of the maximum operational speed, which corresponds to 4~Hz for this dataset.

% \begin{figure}[!htbp]
%     \centering
%     \includegraphics[width=0.8\linewidth]{figures/correctioncagebal3540hzrfs.pdf}
%     \caption{Frequency bands used to define the center and surround receptive fields for the cage fault (left) and ball fault (right). For the cage fault, ten harmonics are visible, while for the ball fault, two sidebands per harmonic can be observed. The bandwidths for the center and surround are 4 Hz and 12 Hz, respectively.}
%     \label{fig:enter-label}
% \end{figure}

%\subsubsection{Signal filtering}
% For implementing the receptive fields, we considered filtering in the time-domain. This is computationally efficient and mathematically sound.
% Using IIR Butterworth filters with different orders controlled by the parameter \emph{Stepness} of bandpass implementation in MATLAB, we experimented with different values and inspected the frequency response of these options. The frequency response plots can be seen in Figure~\ref{fig:butterworth}.

% \begin{figure}[!htbp]
%     \centering
%     \includegraphics[width=0.8\linewidth]{figures/timefilterrfs.pdf}
%     \caption{IIR Butterworth filters of different orders were tested to implement the receptive field operation. The top panel shows the temporal vibration signal, while the bottom panel displays the frequency response plot. The center filter uses a \emph{Steepness} parameter of 0.999, and the surround filter uses a \emph{Steepness} parameter of 0.5.}
%     \label{fig:butterworth}
% \end{figure}

% The second option involves applying the filter in the 
For implementing the receptive fields, we considered filtering in the frequency domain.
This alternative is straightforward and has the advantage of not being restricted by causality, since the entire signal snapshot is available, allowing for greater flexibility in the design of the spectral filter.
Furthermore, operating in the spectral domain enables the use of a Gaussian function as the gain profile, which can be easily shaped to emphasize desired frequency bands.
We will refer to such a filter as a \emph{spectral mask}.

The set $\mathcal{M}$ of \emph{admissible spectral masks} is defined as
$\mathcal{M} = \left\{ m : \mathbb{R}_0^+ \rightarrow [0,1] \right\}$.
We restrict the domain to the non-negative real numbers because the spectral masks will act as a gain that will be multiplied by the magnitude of the frequency components. 
Let $M \subset \mathcal{M}$ be a finite subset of masks. 
We define the disjunction over $M$ as the pointwise maximum over magnitudes as:
\[
    \bigvee M \in \mathcal{M}, \quad \text{specifically}, \quad \bigvee M (f) = \max_{m \in M} m(f).
\]

Given the frequency band $B = [f_{\min}, f_{\max}]$, and the parameter $k_\sigma$, the Gaussian frequency mask $G(f)$ is defined as:

\begin{equation}
    G(f;B,k_\sigma) = \exp\left( -\frac{1}{2} \left( \frac{f - f_c(B)}{\sigma_f} \right)^2 \right)
\end{equation}

with $f_c(B) = \frac{f_{\min} + f_{\max}}{2}$, and $\sigma_f = \frac{f_{\max} - f_{\min}}{2 \cdot k_\sigma}$. The parameter $k_\sigma$, called the sigma rule, determines how the limits of the frequency bands are handled. 
A sigma rule of 3 corresponds to 99.7\% of the area under the Gaussian falling within the specified frequency band. %Figure~\ref{fig:spectral-filtering} summarizes the spectral filtering steps, showing the effect of the filtering in the time domain, although this is not part of our processing pipeline. 

% \begin{figure}[!htbp]
%     \centering
%     \includegraphics[width=0.8\linewidth]{figures/rf-diagram.pdf}
%     \caption{Illustration of the steps followed to compute spectral filtering using the Gaussian frequency mask $G(f;B,k_\sigma)$. 
%     The spectrum of the vibration signals is computed using the Fast Fourier Transform (FFT). 
%     For each center and surround region across the harmonics and sidebands, the frequency mask is multiplied by the signal spectrum to perform selective spectral filtering. The inverse FFT (IFFT) is shown only to illustrate the effect of the spectral filter in the time domain; it is not part of the actual processing pipeline.}
%     \label{fig:spectral-filtering}
% \end{figure}

% \begin{figure}[!htbp]
%     \centering
%     \includegraphics[width=0.8\linewidth]{figures/spectralfiltergaussianfreqtime.pdf}
%     \caption{Spectral filter response for Gaussian frequency masks centered at the characteristic frequencies of outer race faults (10 harmonics). The time-domain signal was reconstructed for demonstration purposes only and is not part of the condition monitoring processing pipeline.}
%     \label{fig:spectral-filtering-time}
% \end{figure}

\subsection{SFRFs Computation}
An advantage of filtering in the frequency domain is that it enables all operations to be computed simultaneously by precomputing a single gain mask across the spectrum for each operational mode. 
This strategy is particularly efficient because, although the characteristic frequencies of interest shift with the shaft speed, the frequency-domain filters can be generated in advance, and applying the filter is equivalent to a Hadamard product (elementwise multiplication) between the spectrum of the vibration signals and the mask corresponding to the appropriate operational mode.

The characteristic frequencies for each fault mode are defined as shown in Table~\ref{fig:characteristic-freqs-faults}. 
We define the corresponding \emph{frequency bands} as $\mathcal{B}(F,W) = \{[f-\frac{W}{2}, f+\frac{W}{2}] \mid f \in F\}$

\begin{table}[ht]
\centering
\renewcommand{\arraystretch}{1.4}
\begin{tabular}{ll}
\hline
\textbf{Fault Mode} & \textbf{Characteristic Frequencies} \\
\hline
Outer race &
$\mathcal{F}_O = \left\{ n f_{\mathrm{BPFO}} \mid n = 1..N_h \right\}$ \\

Inner race &
$\mathcal{F}_I = \left\{ n f_{\mathrm{BPFI}} + s f_r \;\middle|\; n = 1..N_h,\; s = -N_s..N_s \right\}$ \\

Ball&
$\mathcal{F}_B = \left\{ n f_{\mathrm{BSF}} + s f_{\mathrm{FTF}} \;\middle|\; n = 1..N_h,\; s = -N_s..N_s \right\}$ \\

Cage &
$\mathcal{F}_C = \left\{ n f_{\mathrm{FTF}} \mid n = 1..N_h \right\}$ \\
\hline
\end{tabular}
\caption{Characteristic frequencies for bearing faults including harmonics and sidebands. Notation $N_h$: number of harmonics, and $N_s$: number of sidebands}.
\label{fig:characteristic-freqs-faults}
\end{table}

Given the set the characteristic frequencies $\mathcal{F} \in \{ \mathcal{F}_O, \mathcal{F}_I, \mathcal{F}_B, \mathcal{F}_C \}$, and shape parameters $\sigma_C = (\mathcal{W_C},\kappa_C)$ for the center and $\sigma_S = (\mathcal{W_S},\kappa_S)$ for the surround. 
Then given $\sigma = (\mathcal{W},\kappa) \in \{\sigma_C,\sigma_S\}$ we can define a \emph{receptive field gain function} as: 
\[
\mathcal{G}_{\mathcal{F}}^\sigma \in \mathcal{M}, \quad \text{specifically,} \quad \mathcal{G}_{\mathcal{F}}^\sigma  = \bigvee \left\{ G(f; B, \kappa) : 
    B \in \mathcal{B}(F, \mathcal{W}) 
\right\}.
\]
We refer to $\mathcal{W_C}$ as the center bandwidth and $\mathcal{W_S}$ as the surround bandwidth.

\begin{figure}[htb]
    \centering
    \includegraphics[width=0.8\linewidth]{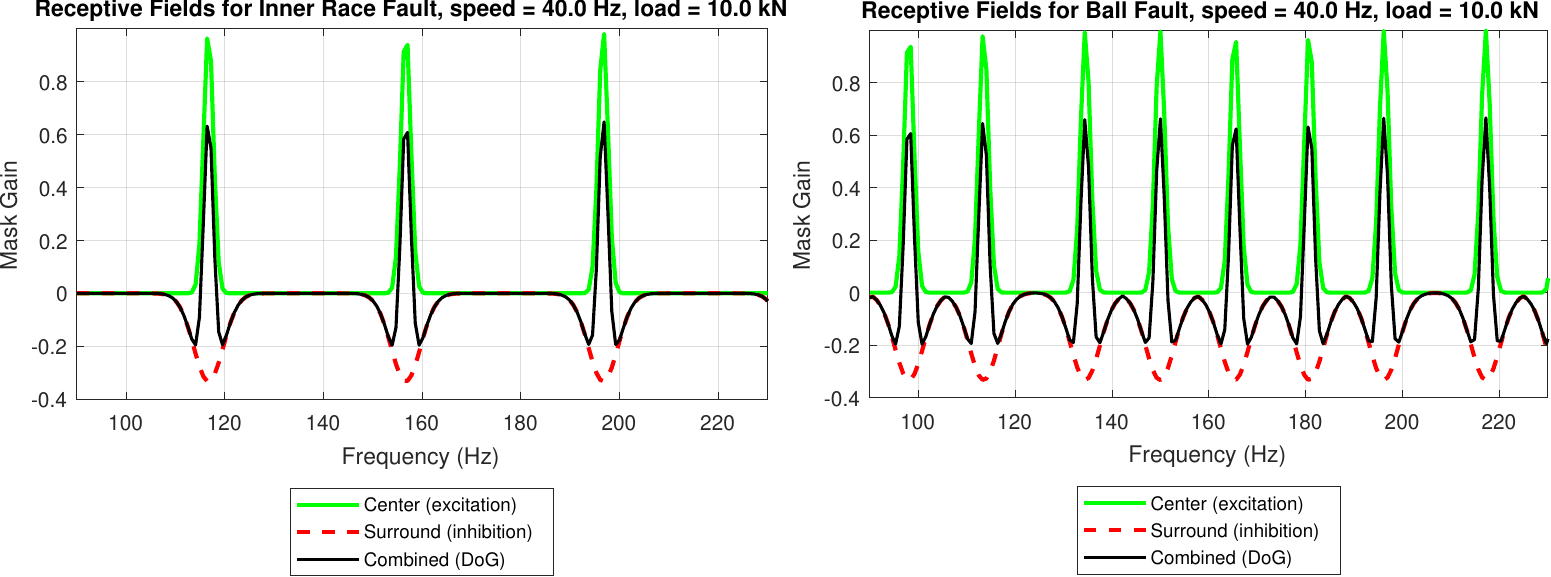}
    \caption{Receptive field visualization for two different fault modes across the frequency spectrum. 
    The DoG parameters are set as follows: $F = \mathcal{F}_I$ for the inner race fault, $F = \mathcal{F}_B$ for the ball fault, $\kappa_C = \kappa_S = 2$, and $\kappa_H = \frac{1}{3}$.
    Left panel: For the inner race fault at 40 Hz shaft speed, the second negative sideband, first negative sideband, and characteristic frequency are visible, from left to right within the panel.
    Right panel: For the ball fault at the same speed, the first positive sideband and second positive sideband of the first harmonic, the second harmonic with all its sidebands, and the second negative sideband for the second harmonic are shown, from left to right within the panel.}
    \label{fig:freq-masks}
\end{figure}

\begin{figure}[p]
    \centering
    \includegraphics[height=0.95\textheight]{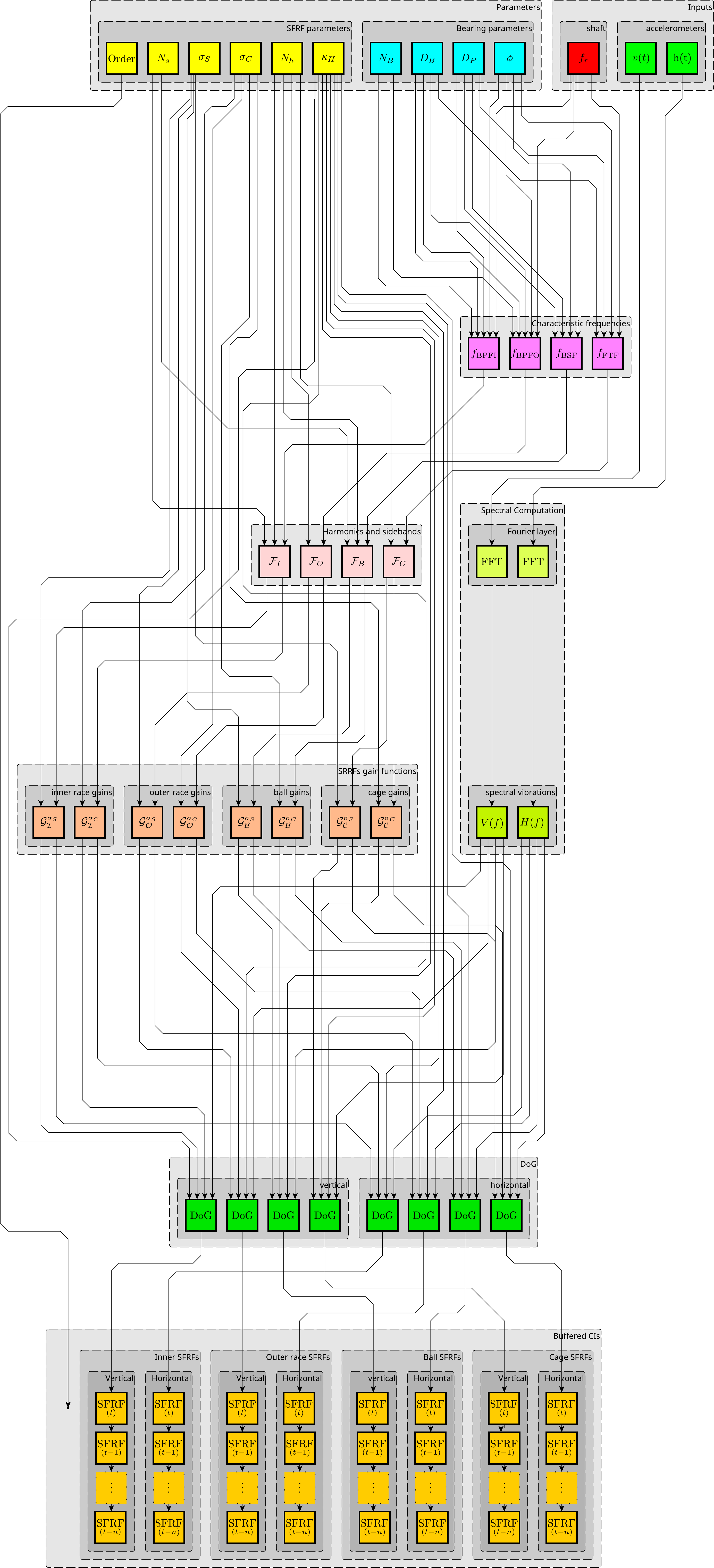}
    \caption{Processing pipeline for the computation of 
    condition indicators using buffered SFRFs.}
    \label{fig:processing-pipeline}
\end{figure}

The \emph{Difference of Gaussians} used to compute the SFRF is then given by:
\begin{equation}
 \operatorname{DoG} =\;
    \int_0^{\frac{f_s}{2}}{ \left[ 
    \mathcal{G}_{\mathcal{F}}^{\sigma_C}(f) - 
    \kappa_H \, \mathcal{G}_{\mathcal{F}}^{\sigma_S}(f) \right]  \left| A(f) \right| \, df}
\label{eq:DoG}   
\end{equation}

where $f$ denotes frequency in the vibration spectrum, $f_s$ the sampling frequency, and $A(f)$ is the Fourier transform of the accelerometer signal. 
The parameters $\kappa_C$ and $\kappa_S$ control the width (sigma rule) of the center and surround Gaussians, respectively, and $\kappa_H$ is the inhibition factor. Figure~\ref{fig:freq-masks} shows the SFRF gain functions and DoG, while Figure~\ref{fig:processing-pipeline} illustrates the SFRF processing pipeline. This pipeline incorporates a memory buffer, whose purpose and order parameter will be detailed later in the paper.

\section{Experimental Evaluation}
In this section, we evaluate the suitability of SFRFs for condition monitoring. 
Traditionally, condition monitoring relies heavily on identifying condition indicators and health indices that effectively capture the degradation trend of a system. 
This effectiveness is often assessed using metrics such as monotonicity, prognosability, and trendability. 
However, the XJTU-SY dataset contains only five samples per operational condition, which is insufficient for meaningful analysis of prognosability and trendability, since these metrics require larger datasets. 
Therefore, we selected \emph{monotonicity} as the primary evaluation criterion. 
This choice is justified because, if a signal is to reliably capture degradation and we assume no regenerative processes, as is typical in the mechanical system under study, there must be a consistent correlation between the values of the condition indicators or health index and the operational time of the machine.
However, it is important to note that expecting perfect monotonicity is unrealistic. Aleatoric uncertainties, unknown inputs, and varying environmental contexts naturally introduce fluctuations into the estimations. 
For this reason, we also consider the \emph{smoothness} of the condition indicator as an additional evaluation metric.

\subsection{Empirical Selection of Parameters}
Although we computed the SFRFs for the entire dataset, we report here only on the qualitative behavior of the condition indicators obtained using SFRFs for the different fault types on the bearing labeled \emph{Bearing1\_1}. 
Since SFRF is a novel technique, there were many unknowns regarding their behavior and the appropriate selection of parameters for their implementation. 
In this section, we present results based on empirical parameter selection for the SFRFs.

Drawing inspiration from the classic understanding of the receptive fields of parvocellular ganglion cells in the primate retina, we chose a center contribution that is stronger than the surround and has a narrower span over the spectrum. 
There are two sets of parameters that control the frequency span of the detectors. 
The first set is given by the bandwidths $\mathcal{W_C}$ and $\mathcal{W_S}$, and the second will be given by how sharply the filter attenuates frequencies outside its limits.

When selecting the first set of parameters, several considerations must be taken into account, with spatial overlap being a primary concern. 
If the bandwidths are too wide, there will be significant overlap between the bands corresponding to multiple faults, as well as between different harmonics or sidebands of the same fault mode. 
Whether such overlap is beneficial remains uncertain. 
Since our solution is biologically inspired by the classical view of visual image formation and aims to mimic, to some extent, the qualitative behavior of chromatic ON channels (or at least their good old-fashioned models), we opted for narrower bandwidths. 
At the same time, we avoided making them too narrow, so as not to miss the natural frequency deviations that are known to occur around the fault characteristic frequencies.

Regarding the frequency resolution of the spectrum, we are limited by the XJTU-SY dataset: snapshots were recorded using time windows of 1.28 seconds, resulting in a maximum frequency resolution of 0.78125 Hz. Since each subsequent snapshot is taken one minute apart, we cannot increase the window size to improve frequency resolution, which would be desirable for mechanical fault diagnosis.

Based on these considerations, we selected $\mathcal{W_C} = 4$ Hz (corresponding to 5 samples within the bandwidth) and $\mathcal{W_S} = 12$ Hz.

The second set of parameters to control the spectral span is the sigma rules for the center and the surround, we opted for  $\kappa_C = \kappa_S = 2$.

\subsection{Evaluation of SFRF with Empirical Parameters}

\begin{figure}[htb]
    \centering
    \includegraphics[width=0.8\linewidth]{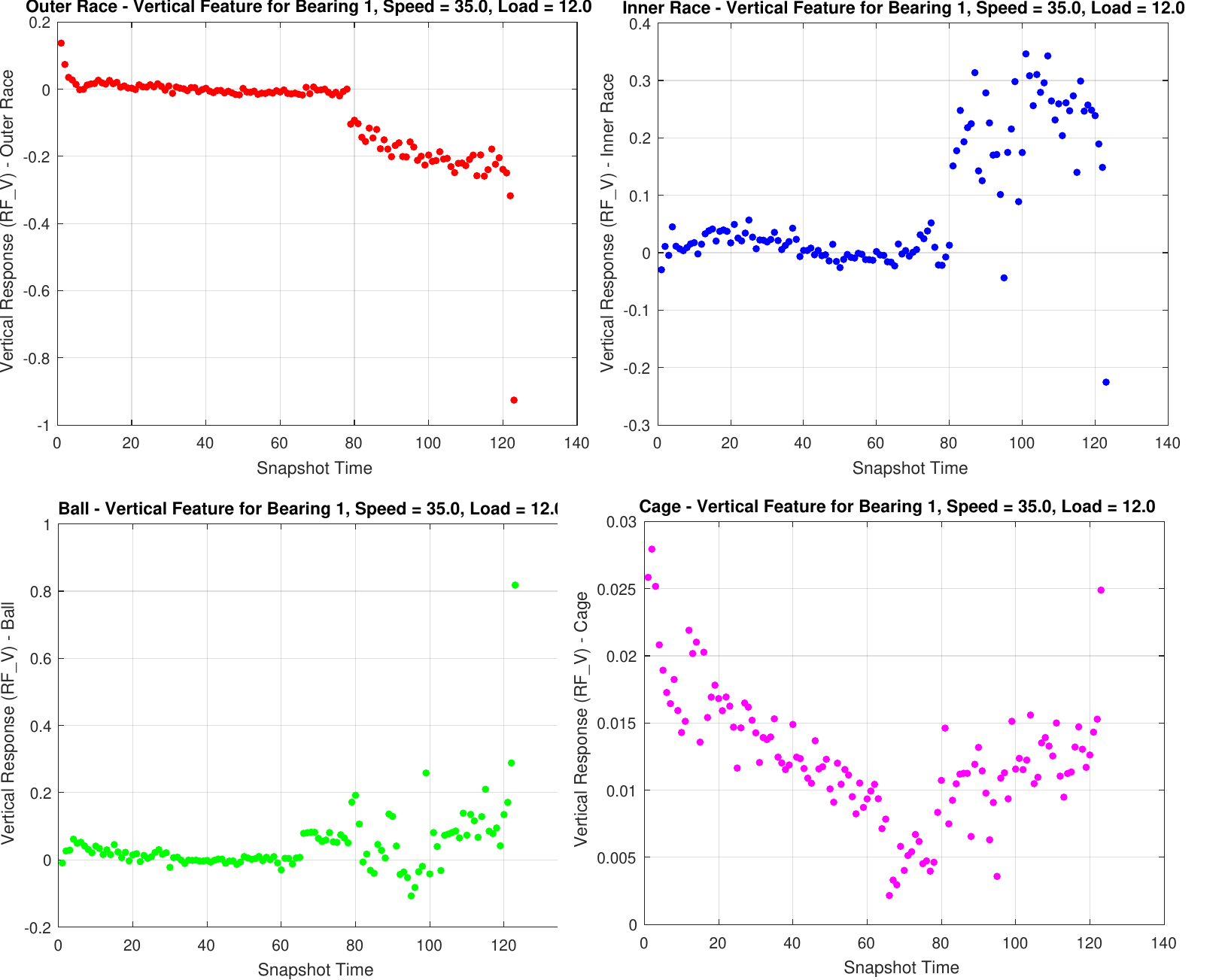}
    \caption{Temporal behavior of SFRFs for the bearing labeled Bearing1\_1. The parameters of the receptive fields were selected empirically, guided by expectations drawn from biological models.}
    \label{fig:empirical-time-vertical-acc}
\end{figure}

We computed the SFRF response to horizontal and vertical acceleration and visualized the temporal behavior of the SFRFs to assess whether they can capture the degradation trend of the bearing.
Figure~\ref{fig:empirical-time-vertical-acc} shows the temporal behavior of the four SFRFs. It can be observed that all SFRFs are capable of detecting a sudden transition in the temporal evolution. This behavior is reasonably interpreted as the manifestation of a defect, with the degradation trajectories for the inner and outer race SFRFs differing significantly around time 80. 
These results support the idea that SFRFs can be used for fault detection and, potentially, for diagnosis, as their outputs are disaggregated according to specific failure modes. 
%However, employing SFRFs for diagnostic purposes would require a consistent and harmonious interpretation of variations in the degradation trends across different failure modes. 
% Figure~\ref{fig:allempirical-vertical} presents these outputs in context with one another, illustrating that they operate at different scales, driven by the relative energy integrated within their respective receptive fields. 
% This observation motivates further exploration into tailoring the SFRFs with independent parameters, as they respond to distinct spatiotemporal signatures associated with different failure modes.

Another notable observation from the ball SFRF in Figure~\ref{fig:empirical-time-vertical-acc} is its early response to an event, around time 65, before outer and inner race SFRFs exhibit any noticeable change. 
This behavior may reasonably be interpreted as indicating either the onset of a less severe defect or a precursor to the severe fault, detected at about time 80 by all SFRFs. 
Notably, this abrupt change in the degradation trajectory is also perceived by the cage SFRF (right-bottom inset). 
The cage SFRF, in particular, captures the early degradation pattern effectively, displaying a consistent trend from the beginning up to the early event. 
This suggests that, even with a crude heuristic parameter selection, the combination of cage and ball SFRFs may offer a reliable monitoring of degradation since the very beginning of the operational life of the bearing.

% \begin{figure}
%     \centering
%     \includegraphics[width=0.8\linewidth]{figures/empiricalall.pdf}
%     \caption{Relative temporal behavior of SFRFs for the bearing labeled Bearing1\_1. The parameters of the receptive fields were selected empirically, guided by expectations drawn from biological models.}
%     \label{fig:allempirical-vertical}
% \end{figure}

% Figure~\ref{fig:3d-degradationtrajectories} shows a 3D visualization of the degradation trajectories for the 4 SFRSs, it can be seen that trajectories are stable but remain responsive to degradation trends.

% \begin{figure}[!htbp]
%     \centering
%     \includegraphics[width=0.8\linewidth]{figures/3ddegradation.pdf}
%     \caption{Degradation trajectories as perceived by the SFRFs in the two dimensions of vibration.}
%     \label{fig:3d-degradationtrajectories}
% \end{figure}

\subsection{Optimizing for Condition Monitoring and Prognosis}

Our qualitative results are encouraging but also highlight several issues with the current method, the most notable being the varying sensitivity of different SFRFs to degradation events. 
Evolutionary multi-objective optimization techniques are particularly well-suited for scenarios where theoretical guidance is limited, as they require minimal assumptions and can efficiently explore complex parameter spaces. 
In this context, we formulated the exploration of the SFRF parameter space as an optimization problem, explicitly quantifying our condition monitoring and prognosis criteria and defining them as objectives to be optimized. 
Table~\ref{tab:objectives} presents these three objectives. 
The first objective directly assesses the model’s predictive accuracy by quantifying the error in remaining useful life estimation. 
The second objective encourages consistent sensitivity to degradation across the component’s lifetime. 
The third objective penalizes jitter along the degradation trajectory, thereby promoting smoother and more interpretable trends.
We chose to use the geometric mean rather than conventional averaging to ensure that no individual SFRF is overlooked during the optimization process. 
Although it is theoretically possible to avoid aggregating the behaviors altogether and instead treat each SFRF as an independent objective, this alternative was not pursued in the present study. 
In retrospect, formulating the optimization of different SFRFs as independent optimization problems may represent a better path, since their computations do not depend on one another.
We leave this possibility open for exploration in future work.

\begin{table}[ht]
    \centering
    \begin{tabular}{l|l}
        Objective & Equation \\
        \hline
        RUL Error (MSE)  &
        $ \frac{1}{K} \sum_{i=1}^K (y^{(t)} - \hat{y}^{(t)})^2$ \\

        Monotonicity     &
        $\left( \prod_{j=1}^F |\rho_j| \right)^{1/F}$  \\

        Smoothness (MAD) &
        $ \left(
        \prod_{j=1}^F 
        \operatorname*{median}\limits_{t}
        \left(
            \left| \Delta x_j^{(t)} - 
                \operatorname*{median}\limits_{t'}
                \left( \Delta x_j^{(t')} \right)
            \right|
        \right)
        \right)^{1/F}$ \\

    \end{tabular}
    \caption{Optimization objectives for NSGA-II. 
    Notation: SFRFs are concatenated into a time-varying condition indicator vector $x^{(t)}$ of dimension F (with 4 SFRFs for horizontal and 4 SFRFs for vertical accelerations), 
     $\rho_j$ is the Spearman correlation between feature $j$ and snapshot time, and $\Delta x_j^{(t)}$ is the first difference of the j-th SFRFs at time $t$, $y^{(t)}$ is the observed RUL at time $t$ while $\hat{y}^{(t)}$ the predicted RUL, both quantities are normalized by the maximum RUL.}
    \label{tab:objectives}
\end{table}

A key advantage of surrogate models lies in the efficient computation of objectives, which is essential given the population-based nature of evolutionary algorithms, where a single run may involve thousands of evaluations.
To predict the remaining useful life (RUL), we trained a bagging regression ensemble model on a sub-sampled degradation trajectory of Bearing1\_1.
This strategy offers several benefits: (1) bagging regressors, as ensemble methods, provide robust predictions with a reduced risk of overfitting; (2) they perform well even with limited data; and (3) they are computationally efficient, making them well-suited as surrogate models for parameter optimization.

However, this decision also introduces a notable challenge: bagging regressors are inherently nondeterministic, which can lead to fluctuations in the Pareto front during optimization.
We consider this acceptable, even somewhat beneficial, as the stochasticity helps mitigate overfitting and discourages convergence toward non-robust regions of the parameter space, which is particularly important when working with limited data.
Further iterations will be necessary to fine-tune key parameters, such as the maximum harmonic and sideband orders.
Nevertheless, premature commitment to computationally expensive studies is not advisable.
Instead, we advocate for an iterative refinement strategy, which facilitates a progressive understanding of both the strengths and limitations of SFRFs as the research evolves.

\begin{figure}[htb]
    \centering
    \includegraphics[width=0.8\linewidth]{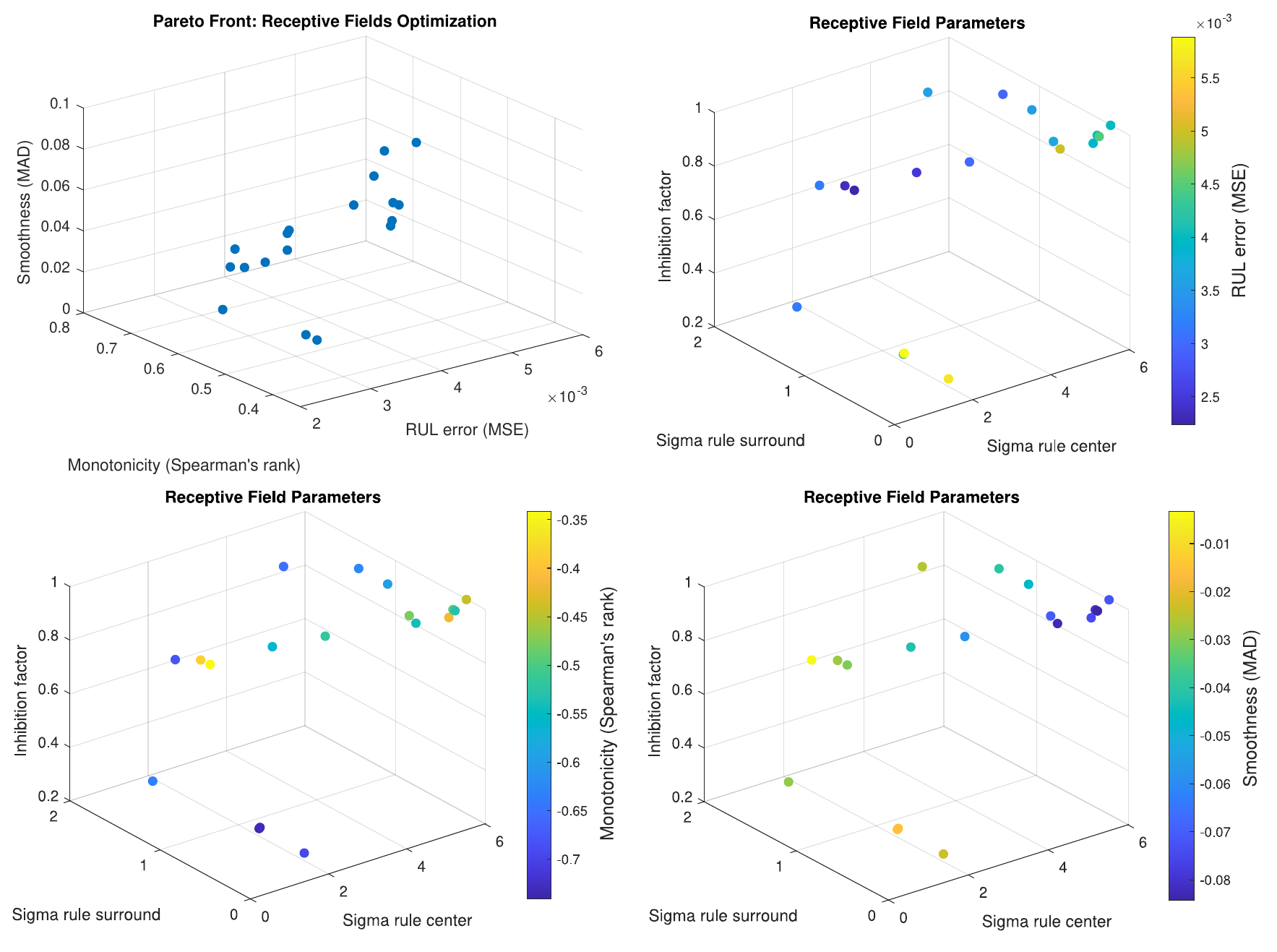}
    \caption{The local Pareto-optimal front identified by NSGA-II after 155 iterations and 7,750 function evaluations is shown.
    In the top left panel, individuals are visualized in the objective space.
    The remaining panels depict the parameter space, with solution colors encoding different objectives. The sign of the monotonicity and smoothness criteria was reversed to enforce their maximization. Parameter space panels:
        top right, RUL estimation error;
        bottom left, monotonicity metric;
        bottom right, smoothness criterion.}
    \label{fig:local-pareto-front}
\end{figure}

We performed the optimization of SFRF parameters with MATLAB's \emph{gamultiobj} function using the algorithm’s default settings and the following domain bounds for the parameters: $\kappa_C, \kappa_S  \in \left[\ n^{-2}, n^{2} \right]\, \kappa_H \in \left[ 0, 1 \right]$ with $n=3$.
Figure~\ref{fig:local-pareto-front} presents the local Pareto-optimal front after 155 iterations, at which point the convergence criterion was satisfied (i.e., the change in the spread of Pareto solutions is less than $1\times10^{-4}$). 
As expected, NSGA-II effectively identifies a diverse set of non-dominated solutions. As the optimization function performs minimization in all objectives, we changed the sign of monotonicity and smoothness to enforce their maximization.

Several findings are noteworthy.
First, all Pareto-optimal individuals cluster within a region where the sigma rule center spreads up to a maximum of 6, while the sigma rule surround spreads up to a maximum of 2. 
This aligns with our expectation that the surround performs better when covering a wider frequency bandwidth (the higher the sigma rule, the stricter the Gaussian). 
However, it is somewhat surprising that this was not already enforced by setting the surround bandwidth $\mathcal{W}_S$ to be three times that of the center $\mathcal{W}_C$. Note that, strictly speaking, $\mathcal{W}_C$ ($\mathcal{W}_S$) can be fused with $\kappa_C$ ($\kappa_S$), but doing so would make their interpretation harder to visualize.

We observe that most objectives conflict with each other. The algorithm’s selection mechanism, which emphasizes boundary individuals through its use of crowding distance, tends to favor solutions at the extremes of the objective space in order to maximize diversity across the Pareto front. 
Despite leveraging interactive visualizations with rotation and projection capabilities, we did not identify the anticipated cooperation between monotonicity and RUL prediction accuracy. 
Specifically, solutions exhibiting high monotonicity (deep blue in the monotonicity inset) often perform poorly in terms of RUL prediction, and vice versa, those with low RUL error (deep blue in the RUL error domain) tend to score low in monotonicity.

This limitation is particularly evident in the case of our cage SFRF indicator, which, despite being highly informative of degradation throughout the entire operational life, would be penalized by conventional monotonicity metrics. 
Its triangular shape, coupled with relatively low energy content and a noisy appearance, would result in a low monotonicity score. 
However, the fundamental issue extends beyond this specific example and lies in the distinction between local and global trends, as well as in the methodology used to compute monotonicity. 
Traditional condition indicators are computed episodically rather than as states of dynamical processes. 
In simple degradation models, the health condition is typically evaluated based on a single temporal snapshot, without consideration for the underlying trend or the temporal evolution of the indicator.
This highlights the need for monotonicity metrics or health indicators that account for temporal dynamics on multiple temporal scales.

These findings suggest that condition indicators should be evaluated as dynamic processes, not merely as isolated episodes. 
This supports the adoption of stochastic differential equations as robust models for degradation processes. 
Moving forward, our research should focus on stochastic model identification and the tracking of their parameters, which may provide a more nuanced and accurate assessment of system health and more rational estimations of remaining useful life.

\begin{figure}[htb]
    \centering
    \includegraphics[width=0.8\linewidth]{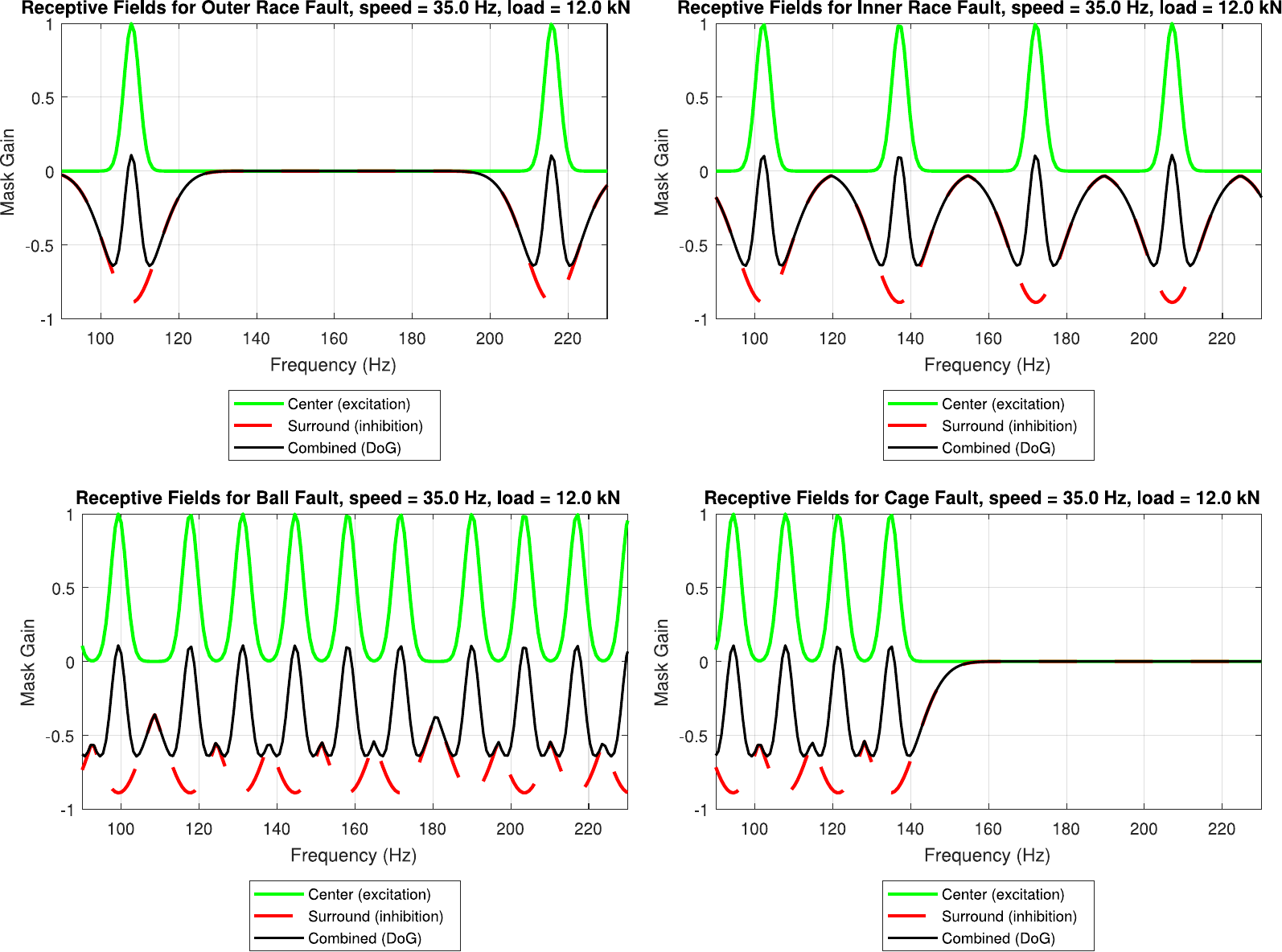}
    \caption{Spectral Fault Receptive Fields (SFRFs) for parameter sets on the local Pareto-optimal front identified using NSGA-II, leading to the most performant RUL prediction via a bagging regressor. Only the range 90-230 Hz band is shown.}
    \label{fig:sfrfs-bestrulpareto}
\end{figure}

We repeated our qualitative evaluation of the SFRFs, this time selecting the individual from the Pareto front that achieved the best RUL prediction performance. This individual is characterized by the following parameters: $\kappa_C = 1.0253$, $\kappa_S = 0.8905$, and $\kappa_H = 0.8647$. Figure~\ref{fig:sfrfs-bestrulpareto} illustrates the SFRFs corresponding to this optimal parameter set within the local Pareto-optimal front, as determined by the lowest RUL prediction error.
The receptive fields exhibit an excitatory center but primarily operate within the inhibitory region.
Notably, a cumulative effect, kept in check by the $\max$ operator, emerges when the Gaussian surrounds overlap, particularly for the ball and cage SFRFs (see bottom insets). This overlap causes the receptive fields to remain continuously sensitive across the local spectrum, thereby achieving spectral (primarily inhibitory) compactness.
As we establish broad parameter ranges for the genetic algorithm to instantiate candidate solutions, we interpret this convergence as indicative of a positive relationship between such spectral sensitivity and the criteria for condition monitoring and prognosis.

\begin{figure}[htb]
    \centering
    \includegraphics[width=0.8\linewidth]{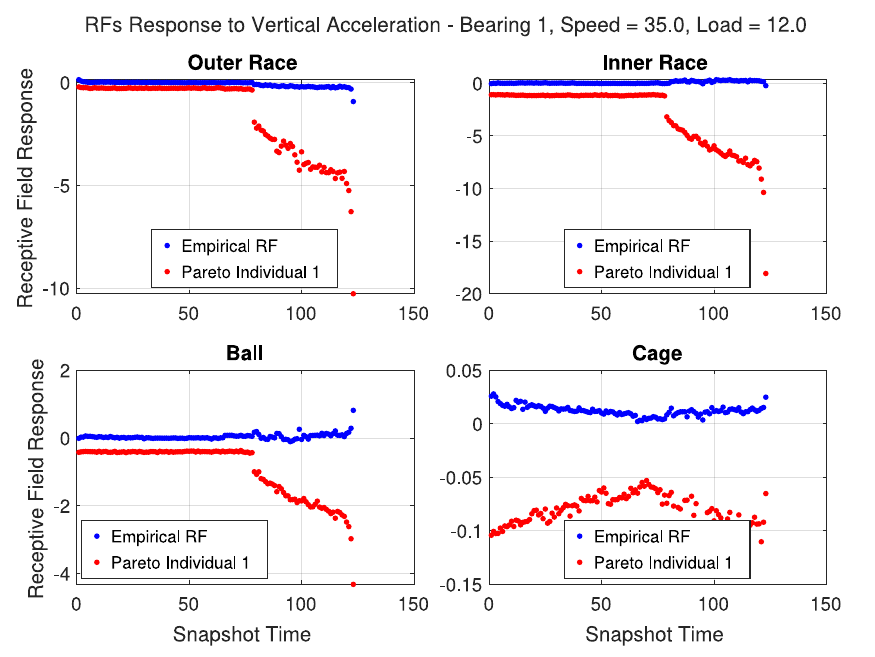}
    \caption{Comparison of condition indicators obtained with empirical parameters versus the best RUL-predicting local Pareto-optimal solution.}
    \label{fig:cis-empirical-vs-bestrulpareto}
\end{figure}

Regarding the utility of SFRFs as condition indicators, Figure~\ref{fig:cis-empirical-vs-bestrulpareto} presents a comparison between the empirically obtained SFRFs analyzed in the previous section and those corresponding to the most performant RUL prediction solution. 
The evolved SFRFs better characterize degradation trends and more clearly signal the onset of defects along the degradation trajectory compared to their empirical counterparts.
Notably, the ability of the cage SFRF to correlate with degradation from the very beginning of the bearing’s operational life is further enhanced by the evolutionary algorithm.

To test our hypothesis that RUL predictions should account for the temporal evolution of condition indicators, we conducted experiments varying the order of the condition indicators used for prediction.
Following standard dynamical systems terminology, we refer to the \emph{zero-order} indicator as the instantaneous condition indicator (although it is computed from a signal snapshot of 1.28 seconds) represented by the eight SFRFs (four fault types across two vibration signals).
The \emph{first-order} indicator is a 16-dimensional vector formed by concatenating the current SFRFs with those from the previous time step.
More generally, the \emph{$n$-th order} SFRF corresponds to an $8(n+1)$-dimensional vector comprising the current SFRFs and a buffer containing the previous $n$ sets of SFRFs.
This formulation allows the model to incorporate temporal context and memory into the RUL prediction process.

\begin{figure}[htb]
    \centering
    \includegraphics[width=0.8\linewidth]{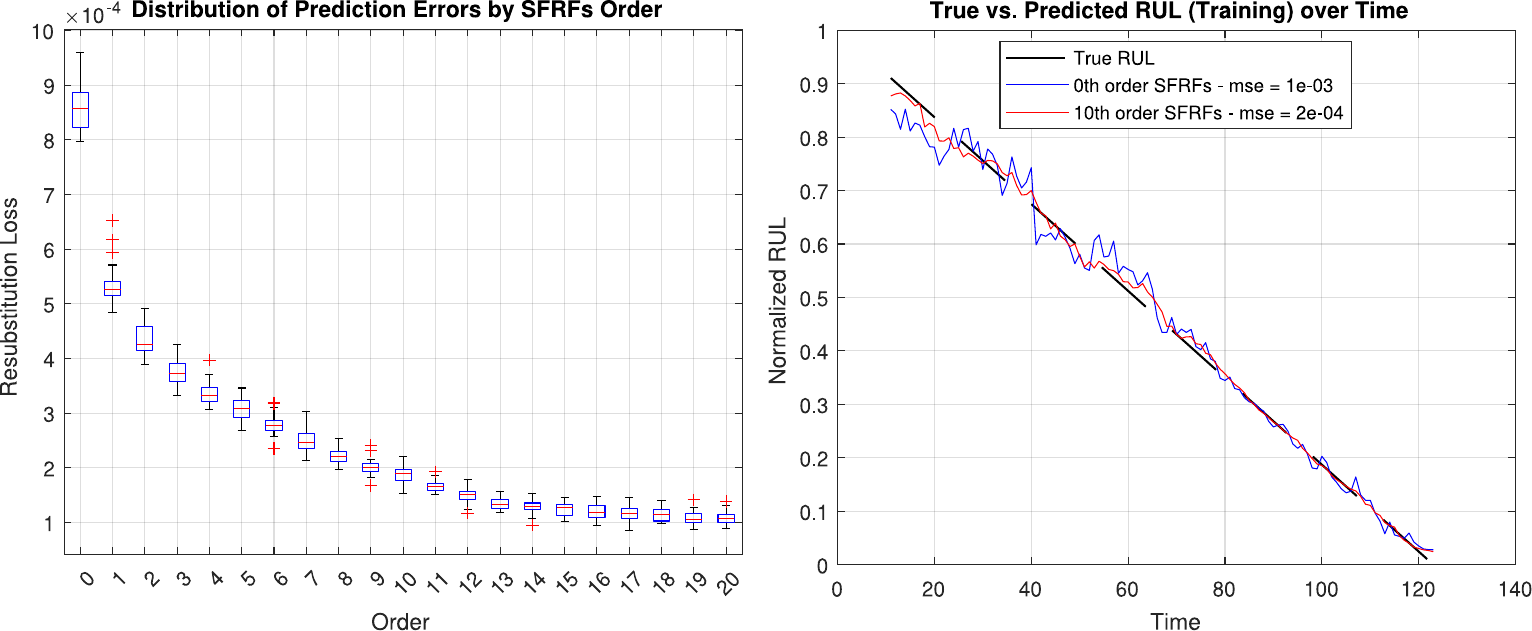}
    \caption{Training (resubstitution) MSE error with increasing orders and comparison of RUL estimations between 0th order and 10th order SFRFs.}
    \label{fig:predictions-training-orders}
\end{figure}

Figure~\ref{fig:predictions-training-orders} illustrates the effect of varying the SFRF order on prediction performance.
The left inset shows the resubstitution loss of bagging regressor models trained with different orders.
Due to the model's nondeterministic nature, we repeated the training 30 times and used box plots to represent the distribution of errors for each order.
We observe that using a second-order SFRF condition indicator vector can reduce the resubstitution loss by approximately half.
The right-hand visualization demonstrates the impact of SFRF order on RUL prediction accuracy; notably, the 10th-order predictor exhibits a marked improvement, closely tracking the true RUL across the entire operational life.

It is worth noting that, since signal snapshots are recorded at one-minute intervals, employing a 10th-order SFRF implies buffering SFRF computations from up to 10 minutes in the past, requiring the simultaneous storage of $(10 +1) \times 8 = 88$ floating-point values in memory.

While these results are encouraging, they reflect only the \emph{training} loss and must be substantiated through rigorous cross-validation methodologies.
Nevertheless, the findings underscore the potential value of incorporating temporal memory into RUL estimation.
We will design corresponding validation experiments by holistically interpreting the qualitative results presented in this work, ensuring that all observed patterns and trends are appropriately factored into the evaluation.

\section{Discussion}

Drawing inspiration from the biology of vision, we have explored the implementation of SFRFs based on the characteristic frequencies of bearing elements, their harmonics, and known amplitude modulation phenomena. 
SFRFs compute spectral activity contrast in a manner analogous to how visual systems encode chromatic information. 
However, instead of excitation by light and preferential responses to different wavelengths by photopigments in cone photoreceptors, we analyze vibration signals transformed into the frequency domain. 
By monitoring the characteristic frequencies associated with different bearing components, SFRFs enable us to track degradation trends throughout the operational life of the machinery.
Our particular implementation relies on the computation of Gaussian spectral filters centered at the characteristic frequencies, numerical integration of frequency bins across the spectrum, and the evaluation of spectral contrast, which depends on integrated energy within a narrow bandwidth we call the center, and a wider bandwidth, the surround. 
The DoG model, adapted to our domain, was formally defined and implemented. 
Its definition aims for computational efficiency.
Our qualitative evaluation of the SFRFs for monitoring health state demonstrates their potential as effective condition indicators. 
The trends observed throughout the operational history indicate that SFRFs can detect abrupt defect events. 
Furthermore, they appear capable of capturing the gradual evolution of degradation. 
Our findings suggest that different SFRFs exhibit unique behaviors and require appropriate parameter tuning to maximize their effectiveness. 
Building on these insights, we established quantitative criteria to assess the suitability of various parameters in the Difference of Gaussians (DoG) model, ensuring the extraction of features that are most relevant for condition monitoring and prognosis.

We use a multiobjective genetic algorithm to compute an approximate Pareto-optimal set of solutions. To guide the optimization, we incorporated three objectives: first, the RUL prediction error, evaluated using fast surrogate RUL estimators instantiated as bagging regressor models; second, monotonicity, which is widely recognized in the PHM community as a crucial metric for feature selection in prognostic pipelines; and third, smoothness, which targets the desirable property of stability in condition monitoring indicators. 
We contrasted the quality of the best predictor against the empirical counterpart, demonstrating the value of the optimization stage.
We also observed, through analysis of the cage SFRF, that certain feature indicators may provide valuable information for condition monitoring and prognosis, yet may be overlooked or rejected when evaluated solely by standard monotonicity metrics. 
This highlights the need for the development of more sophisticated evaluation criteria that can capture the full prognostic value of such features. 
Motivated by this observation, we investigated the impact of incorporating local temporal trends by stacking the SFRFs in a memory buffer. 
We assessed RUL predictions in episodic estimations across different orders, that is number of time samples in the buffer, and our results confirm that incorporating temporal context significantly reduces prediction error.

We acknowledge the preliminary nature of our contribution. 
Notably, the distinct spectrotemporal properties exhibited by different SFRFs suggest that their parameters should be tuned independently, as a one-size-fits-all approach may be overly simplistic. 
Drawing further inspiration from biology, it is well established that retinal ganglion cells operate in parallel channels, each capturing diverse spectro-spatio-temporal properties and contributing to a robust and flexible representation. 
Similarly, future research could explore the simultaneous deployment of a diversity of solutions along the Pareto front through ensemble techniques. 
By orchestrating and interpreting the responses of multiple SFRFs, it may be possible to achieve more comprehensive and adaptable representations, where individual SFRFs provide complementary, partial views tailored to specific objectives. 
For instance, some SFRFs may be better suited for condition monitoring, others for RUL prediction, and their relative importance or activity could adapt dynamically depending on the degradation state.
Currently, FFT computation is the most resource-intensive step in the pipeline. However, next-generation sensors could be designed to shift the focus from general-purpose accelerometers to resonant arrays that respond preferentially to the engine’s spectral fingerprint, potentially eliminating the need for FFT altogether.

\section{Conclusions}
This study demonstrates the value of drawing inspiration from nature to develop robust and reliable systems. 
Spectral fault receptive fields show considerable promise as foundational elements for condition monitoring and prognosis. 
They also have a minimal computational footprint, making them well-suited for onboard deployment in EVPs.
Our qualitative evaluation indicates that, particularly in their optimized form, SFRFs are well-suited for both condition monitoring and remaining useful life (RUL) estimation.

We conclude that by integrating established vibrational analysis techniques with conceptual models from biological perception, and with the help of evolutionary algorithms, it is possible to devise effective solutions for tracking degradation states throughout the operational life of bearings. 
These types of biologically inspired solutions open new possibilities for advancing predictive maintenance and enhancing the reliability of industrial machinery.

\section*{Supplementary Material}
The MATLAB notebook supporting this paper is available at \href{https://doi.org/10.5281/zenodo.15660819}{doi:10.5281/zenodo.15660819}.

\section*{Acknowledgments}
SMG thanks Mike Denham for introducing him to the Biology of Vision and Ian Parmee for introducing him to Evolutionary Multi-Objective Engineering Design.

\section*{Funding}

This study was conducted within the framework of the project ARCHIMEDES, receiving funding from the Key Digital Technologies Joint Undertaking (KDT JU) - the Public-Private Partnership for research, development, and innovation under Horizon Europe – and National Authorities under Grant Agreement No 101112295 funded by the European Union and the FFG under Grant Agreement No FO999899377.

\bibliography{sfrfs}

\end{document}